\pdfoutput=1

\documentclass[11pt]{article}

\usepackage{ACL2023}

\usepackage{times}
\usepackage{latexsym}
\usepackage{graphicx}
\usepackage[T1]{fontenc}
\usepackage[utf8]{inputenc}

\usepackage[T1]{fontenc}

\usepackage[utf8]{inputenc}

\usepackage{microtype}

\usepackage{inconsolata}
\usepackage{graphicx}
\usepackage{natbib}

\title{\textbf{HonkaiChat: Companions from Anime that feel alive!} \\ \small{Event driven conversations for engaging companionship in fantasy role-playing}}

\author{
  Shilong Guo\textsuperscript{1}, Yueze Liu\textsuperscript{1,2,*}, Yichi Zhang\textsuperscript{1}, Zhaoyang Zhu\textsuperscript{1}, Shaan Om Patel\textsuperscript{2} \\
  \textsuperscript{1}University of Illinois at Urbana-Champaign (UIUC) \\
  \textsuperscript{2}Divergence 2\% Research Laboratory \\
  \texttt{yuezel2@illinois.edu} \\
}
\date{\today}

\begin{document}
\maketitle

\renewcommand{\thefootnote}{\fnsymbol{footnote}}
\footnotetext[1]{Corresponding author.}
\renewcommand{\thefootnote}{\arabic{footnote}}
\begin{abstract}
Modern conversational agents, including anime-themed chatbots, are frequently reactive and personality-driven but fail to capture the dynamic nature of human interactions. We propose an event-driven dialogue framework to address these limitations by embedding dynamic events in conversation prompts and fine-tuning models on character-specific data. Evaluations on GPT-4 and comparisons with industry-leading baselines demonstrate that event-driven prompts significantly improve conversational engagement and naturalness while reducing hallucinations. This paper explores the application of this approach in creating lifelike chatbot interactions within the context of Honkai: Star Rail, showcasing the potential for dynamic event-based systems to transform role-playing and interactive dialogue.
\end{abstract}
\section{Introduction}
Modern conversational agents, including anime-themed chatbots, are frequently reactive and personality-driven. They rely on carefully tuned prompt instructions to remain "in-character," responding to user queries in a way that reflects a set personality, background, or narrative role. However, these systems often fail to capture the dynamism of everyday human interactions, which are shaped not only by immediate exchanges but also by evolving events, contexts, and moods. For instance, a conversation might be influenced by earlier events such as a fire alarm at work, discovering a new skill, or admiring a beautiful flower. These factors shift moods, alter perspectives, and influence conversational topics. Despite these nuances, current chatbot systems rarely incorporate such “life-like” elements, operating instead as static, context-reactive agents without an evolving internal state.
Existing frameworks attempt to address conversational depth in distinct ways. For example, CharacterLLM\cite{shao2023characterllmtrainableagentroleplaying} aligns a personality more faithfully by using interview-style questions and refining responses to adhere to specific traits, though it remains largely reactive to user prompts. Retrieval-Augmented Generation (RAG)\cite{lewis2020retrieval} pipelines incorporate past conversation logs or external knowledge to contextualize responses, which improves grounding but lacks proactive narrative development or a sense of internal “life.” Both approaches result in user-driven, reactive systems, leaving an unmet demand for bots that feel "alive"—bots with motivations, event-driven states, and evolving moods.
To address this gap, we explored the use of event-driven prompts to enhance conversational depth and engagement. Our approach was tested against industry-leading models, including Character.AI, where we used carefully engineered prompts embedding dynamic events to evaluate their effect on conversational quality. The case study involved GPT-4 as an evaluator, rating conversations between two random characters interacting with a lead character. The results consistently demonstrated that introducing events significantly enhanced engagement, making the interactions feel more natural and immersive.
We also applied this event-driven framework to a chatbot based on the popular mobile game Honkai: Star Rail by miHoYo, which has been widely embraced by young people across multiple regions. In particular, we brought the character March 7th to life, enabling her to interact with users in a more human-like manner. To achieve this, we crawled text and dialogue data from sources like the Honkai: Star Rail Wiki and pre-trained a Llama 3.1 8b model on the corpus to instill universal knowledge of the game. Subsequently, we fine-tuned the model with curated dialogues and character-specific events to shape March 7th's personality and responses.
Our hypothesis is that seeding conversational agents with rich, character-specific events and training them to respond in a character-consistent manner leads to more engaging, varied, and believable interactions. The experimental results demonstrated that event-driven prompts provide a clear improvement in conversational quality compared to static or context-reactive systems. This work highlights the potential of event-based systems to elevate the immersive experience of role-playing chatbots, making fantasy roleplaying feel more at home by bridging the gap between static responses and the nuanced, evolving nature of human interactions.

\section{Related Work}
\subsection{CharacterLLM}
CharacterLLM proposes a framework of trainable agents for role-playing which simulates the role by instilling his/her experiences, characteristics, and emotions\cite{shao2023characterllmtrainableagentroleplaying}. The model is firstly trained with the memories flashes which is constructed by the character’s profile data, followed by a fine-tune stage which instill the model of contextual information of the characters. The main drawback of this framework is that the role is trained based on profile data which cannot sufficiently represent the whole life and context of the person. Another drawback of the strategy used in CharacterLLM experiments is that the model’s performance is degraded by the vanilla base model whose pre-training data range in distribution does not align with the role-specific data. 

\subsection{Retrieval Augmented Generation}
RAG\cite{lewis2020retrieval} architecture consists of a retriever and a generation-oriented LLM. It helps to reduce the problem of hallucinations when using pure LLM for text generation by combining the information retrieved from a third source, such as documents, external data sources, etc. It provides the generation model a hook to fetch domain-specific data to improve the quality of prompts. In the application of a real-time chatbot, the retriever could be used to fetch relevant background information and enable the bot to generate more accurate and event-focusing responses. The key drawback of RAG is it takes a longer cost than a pure generation model due to the retrieval process, and the model may produce odd responses if the retrieval results do not fully align with the current events.

\subsection{ChatHaruhi}
ChatHaruhi\cite{li2023chatharuhi} proposed a complete role-playing algorithm system to play real characters from anime or TV during a conversation by training a LLM to learn the background knowledge, the personality of the character, and the character’s linguistic habits. The key idea is to extract as much of the original script as possible to form the base memory for a character and search relevant plots to form a prompt when users input a question. The main pipeline of our project is inspired by this system.

\section{Data}
Our dataset is centered around the popular game Honkai: Star Rail, a title that has gained significant traction within the role-playing community. This dataset serves as the foundation for testing and enhancing our prompt augmentation methodology.

\subsection{Data Creation and Augmentation}
To demonstrate our approach, we tasked GPT with generating scenarios that were “casual, surprising, and realistic,” ultimately producing and refining 50 scenarios per character across 26 characters, resulting in 1,300 data points. The process required extensive grounding in the game’s lore and knowledge base, making it a labor-intensive step. Additionally, we incorporated detailed OCEAN and MBTI personality analyses to better support role-playing and event generation tasks.
For training the role-playing model specifically tailored to the character "March 7th," we focused on creating a dialogue dataset. Due to the novelty of the character, there was a lack of pre-existing training texts. To overcome this, we curated a combination of in-game dialogues and synthesized content. The following sections detail the dataset's sources, collection methods, and preparation.

\subsection{Data Sources}
The primary data source was the Honkai: Star Rail Wiki, a fan-maintained repository featuring comprehensive lore, character descriptions, and dialogue excerpts. We systematically extracted relevant information from the Wiki, focusing on all textual content related to March 7th. Additional sources included in-game dialogues (approximately 4,000 lines), which served as the foundation for generating synthetic conversations.
To address the scarcity of domain-specific content, we used GPT to generate synthetic dialogues. These conversations were grounded in Honkai: Star Rail's narrative universe, ensuring thematic and contextual consistency. Prompts encouraged GPT to simulate discussions about newly created events or lore, significantly enriching the training dataset.

\subsection{Data Augmentation}
To expand and refine the dataset, we implemented the following preprocessing techniques:

\begin{itemize}
    \item Character Name Masking: Dialogues involving multiple characters were altered by masking one character's name (excluding March 7th) to simulate user input. This allowed the model to generate context-appropriate responses for March 7th, enhancing its conversational adaptability.
    \item Noise Removal: Non-dialogue artifacts, such as page formatting remnants or unrelated lore, were filtered out to maintain dataset quality.
    \item Dialogue Synthesis: Building upon the methodology from the CharacterLLM paper, we used GPT-4 to extend existing in-game dialogues. This approach generated new, thematically consistent conversations that adhered to Honkai: Star Rail's narrative and character profiles. By increasing the volume and diversity of training data, this process addressed the challenges of limited domain-specific dialogue data in gaming contexts.
\end{itemize}
\section{Method}
To address the issue of conversation initiation, we propose fine-tuning language models specifically for conversation starters. While traditional LLMs ensure chatbot responses are aligned with the provided text, they often lack a sense of autonomy, as chatbots should exhibit their own thought processes and life experiences that influence their behavior in short conversations\cite{li2023chatharuhirevivinganimecharacter}. Instead of simply providing a fixed setting and immediate context, we suggest incorporating life experiences and randomly generated events that the agent might reference during dialogue, much like a natural conversation.

\subsection{Evaluation of Need}
\begin{figure}[h]
    \centering
    \includegraphics[width=1.0\linewidth]{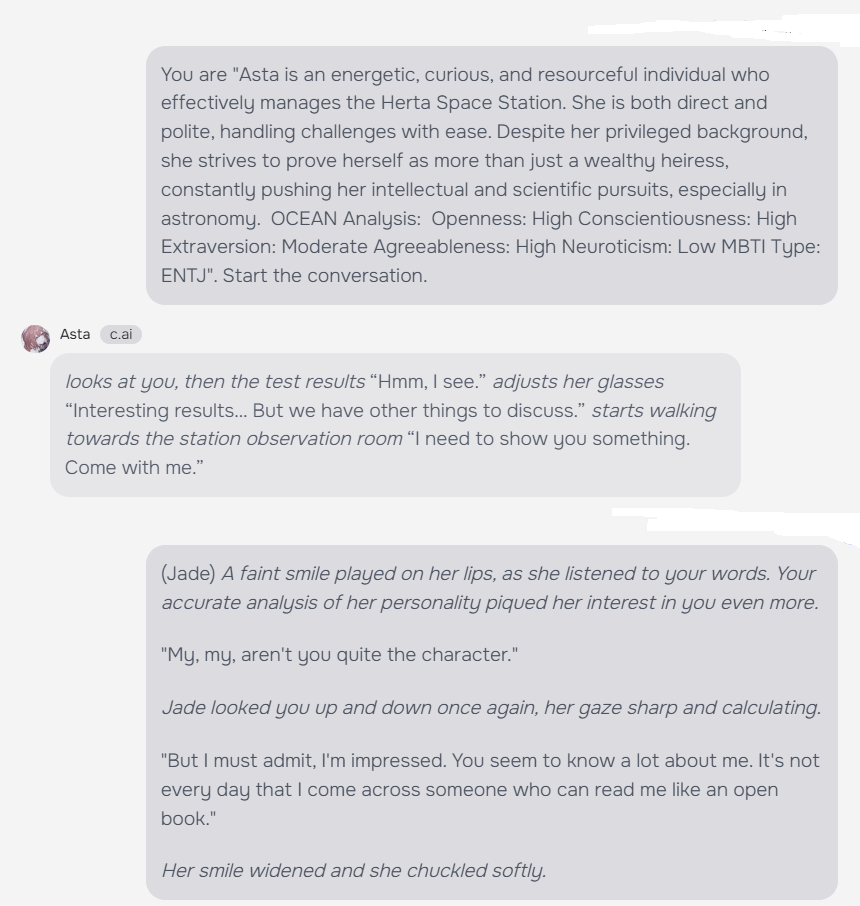}
    \caption{Conversation example starter and reply}
    \label{fig:example1}
\end{figure}
To test out the effectiveness of our method of prompting, we pitted conversation.ai agents with each other. In one case, they were given an event that happened during their day, and in another case, they were not given anything in initializing the conversation the events were randomly chosen from dataset (1). The interestingness is later given to GPT4 for a multi-dimension metric analysis

\subsection{Training our own bot}
Dissatisfied with character AI even with the implementation, we looked to see if we can add our own spin on creating a role-playing agent. We attempted this through 4 stages, though incomplete. The pre-train stage to read 17 MB of data, then fine-tune it on only one character's dialogue to attempt to mimic that person's behaviors. 
\subsection{Training Details}
We constructed 2,500 training pairs based on the existing in-game dialogue dataset from Honkai: Star Rail. Each input includes background information introducing the dialogue scenario and the conversation context, while the output is a standard character response. Additionally, we generated 2,500 more training pairs following characterLLM's approach of dialogue data augmentation. We trained our model on a total of 5,000 input samples, using LLaMA 3.1-8B as the base model. To achieve better performance while considering computational cost constraints, we employed half-precision floating-point calculations, LoRA model\cite{DBLP:journals/corr/abs-2106-09685} based on the PEFT library, and stage 3 zero\_optimization. We used a batch size of 128 and a learning rate of 7e-5. For each character model, the training was conducted on a single Nvidia A6000 GPU and took approximately 2 hours to complete.

\subsubsection{Retrieval-Augmented Generation}
We attempted to use RAG for training, though the integration was deemed unnecessary as we couldn't reach the inference stage where this was relevant. 

\subsection{Data Construction}
\subsubsection{Event Generation}
We first examined the native capabilities of large language models (LLMs) to produce events that could be used as conversation starters. Our preliminary tests revealed that off-the-shelf event generation is often generic and uninteresting. To overcome this, we manually curated a dataset of 1,300 events—about 50 for each of several characters—ensuring that these events varied in scope, emotional tone, and narrative impact(dataset 1). Examples include a character discovering a new coffee-making technique, encountering a mysterious artifact at work, or observing a rare flower.

\subsubsection{Character Reaction Data}
To ground the model in a rich fictional universe, we scraped the Honkai: Star Rail wiki for detailed character backgrounds, lore, and dialogue lines(dataset 2). This provided the model with context about characters’ personalities, histories, and thematic elements. We also supplemented the limited in-game dialogues (roughly 4,000 lines) with GPT-generated conversations anchored in the Honkai: Star Rail universe(dataset 3). These synthetic dialogues were created by prompting GPT to imagine how characters would discuss newly generated events or lore, thus expanding our training corpus.

\section{Evaluation \& Metrics}
In order to evaluate the performance of our model, we conducted the evaluation on a curated subset of in-game dialogue text data, specifically a test split consisting of 200 independent samples (N = 200). For comparison purposes, we introduced LLaMA 3.1-8b-instruct with a prompt-based approach as a baseline model. We leveraged a state-of-the-art, proprietary large language model, GPT-4 (version released on November 20, 2024, referred to as "gpt-4o-2024-11-20"), to perform qualitative evaluations of the generated responses. Specifically, for each data sample, the evaluation setup included the original context, the corresponding ground-truth response from the dataset, the response generated by the baseline (LLaMA 3.1 Instruct), and the response generated by our proposed approach. GPT-4 was tasked with rating the quality of each response across five distinct dimensions from 0-10:

\begin{itemize}

 \item \textbf{ Memorization:} The model’s ability to recall relevant information about the character being portrayed, including precise and detailed knowledge about people, events, and objects associated with the role.

\item \textbf{Values:} The model must share the same objectives and values as the character it portrays, and possesses a distinctive framework for evaluating situations based on the character's perspective, which reflects the character's preferences and biases.

\item \textbf{Personality:} The model should mimic the way that the character would think or speak, such as the speaking style or the tones, and the emotions and reactions under different circumstances.

\item \textbf{Hallucination:} To maintain believability, it is crucial to assess the model's ability to discard knowledge and skills that the character would not have. For example, when questioning an ancient individual about computers, the character should express a lack of knowledge rather than discussing the advantages of modern technology.

\item \textbf{Stability:} Models can be brittle to the influence of pre-training or alignment during prolonged periods of acting, resulting in deviations from the intended portrayal. Our objective is to assess the agent's stability and consistency over a relatively long duration, unaffected by variations in incremental inputs.
\end{itemize}

\subsection{Result}
\begin{figure}[h]
    \centering
    \includegraphics[width=1.3\linewidth]{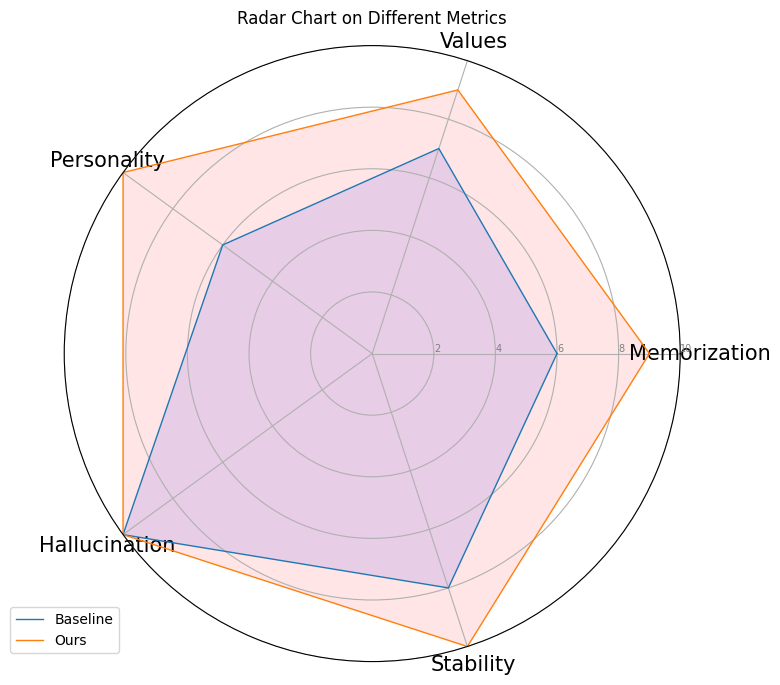}
    \caption{The evaluation result of our model vs base model}
    \label{fig:example4}
\end{figure}
\subsubsection{Event driven prompt Engineering}
The difference between event-driven dialogue and casual conversation was not particularly pronounced, partly because GPT assigned similar scores to most bot conversations. This indicates that, at first glance, the systems lack noticeable variability for users. However, we observed that when no prompt was provided, the bots often defaulted to starting with “may I ask a question.” While acceptable in casual conversations, this repetitive opening could become monotonous over multiple sessions due to its lack of focus on specific topics. Our re-alignment though doesn't immediately increase sensitivity, provides a framework for generic long-term interactions.
\begin{figure}[h]
    \centering
    \includegraphics[width=1.0\linewidth]{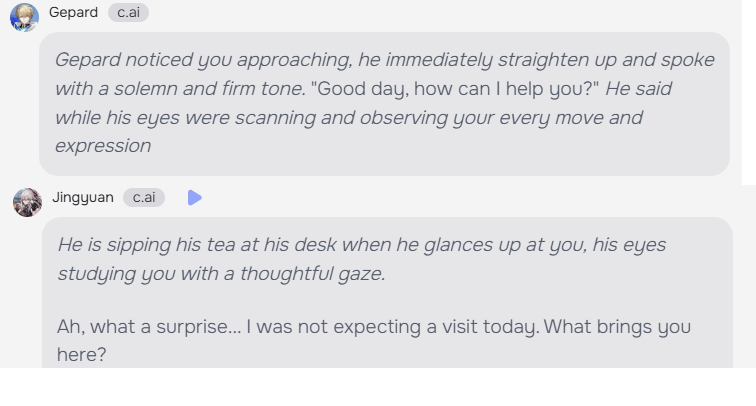}
    \caption{A typical conversation starter for interacting with LLMs}
    \label{fig:example2}
\end{figure}

\begin{figure}[h]
    \centering
    \includegraphics[width=0.8\linewidth]{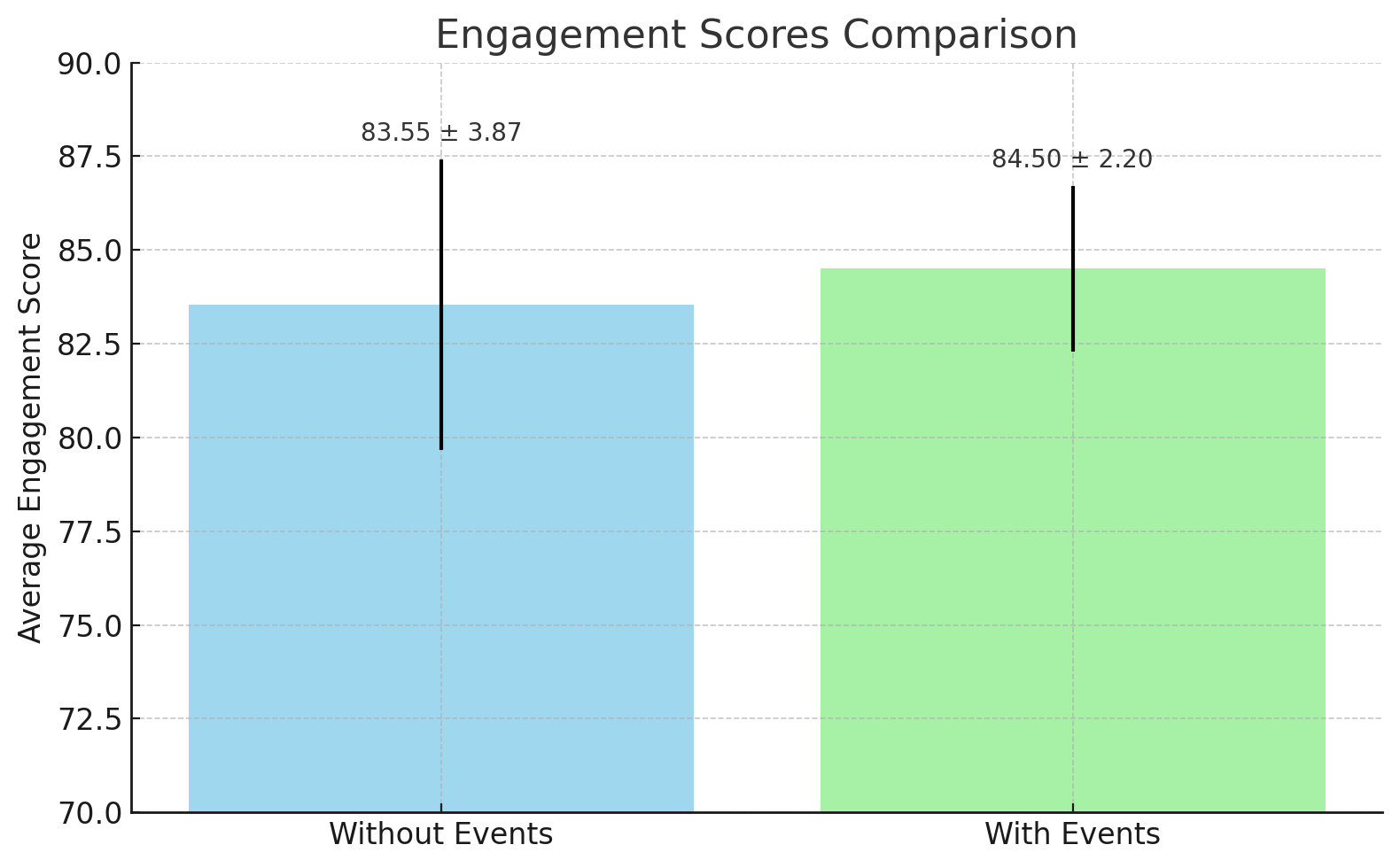}
    \caption{The scores of conversations for engagement for with and without events }
    \label{fig:example3}
\end{figure}

\section{Analysis}
\subsection{Reduced Generic Hallucinations}
When asked about an “imaginary tree” in the game, the model no longer fabricated irrelevant details (e.g., “a tree in London”). Instead, it referenced the in-game concept of “imaginary force,” aligning more closely with the Honkai: Star Rail universe.
\subsection{Contextual Responses to Events}
When prompted with recent events (e.g., “You scored an important contract with a client.”), the model showed more appropriate emotional responses, such as excitement or pride. This indicates some early success in event-driven mood alignment.

\section{Future Works}
\subsection{Maintaining Conversational Focus}
Long inputs with extensive character analysis sometimes caused the model to shift from conversation to a narrative storytelling mode. This suggests we need better prompt strategies or architectural adjustments to keep the model in a conversation-oriented state.

\subsection{Alignment and Personality Diversity}
The model tends to produce “good enough” reactions that reflect generic human behavior. We aspire to achieve more distinct, character-specific reactions. This likely requires further refinement of training data, improved fine-tuning techniques, and possibly reinforcement learning from human feedback to shape nuanced personality traits.

\subsection{Manual tuning}
Our current setup requires manual event inspection for alignment, and our pipeline is also designed to handle one character at a time. Future steps includes fine-tuning a bot that provides proficient event pipelines as well as a one-step framework from text to LLM output pipeline.

\section{Conclusion}
Our preliminary experiments show that incorporating event generation and event-driven reaction patterns into chatbot design can yield more dynamic and contextually interesting conversations. While early results are promising—reducing irrelevant hallucinations and producing more event-sensitive dialogue—significant work remains. We need better alignment, deeper personality imprinting, and more robust conversational grounding. Further iterations will focus on refining event sets, improving fine-tuning strategies, and employing user feedback to shape more authentic, character-specific conversational agents. In the future, this kind of agents are possible to be integrated as monetary feature to be incorporated in to a game as player attachment to fictional characters arise.

\bibliographystyle{abbrvnat}
\bibliography{custom}
\end{document}